\documentclass[a4paper,twoside]{article}

\usepackage[hidelinks]{hyperref}
\usepackage{booktabs}

\usepackage{epsfig}
\usepackage{subcaption}
\usepackage{calc}
\usepackage{amssymb}
\usepackage{amstext}
\usepackage{amsmath}
\usepackage{amsthm}
\usepackage{multicol}
\usepackage{pslatex}
\usepackage{apalike}
\usepackage[bottom]{footmisc}
\usepackage{SCITEPRESS}     

\begin{document}

\title{Salient Mask-Guided Vision Transformer for Fine-Grained Classification}

\author{\authorname{Dmitry Demidov,
Muhammad Hamza Sharif,
Aliakbar Abdurahimov,
Hisham Cholakkal
and Fahad Shahbaz Khan
}
\affiliation{
Mohamed bin Zayed University of Artificial Intelligence, Abu Dhabi, UAE}
\email{\{dmitry.demidov, muhammad.sharif, aliakbar.abdurahimov, hisham.cholakkal, fahad.khan\}@mbzuai.ac.ae}
}


\abstract{
    Fine-grained visual classification (FGVC) is a challenging computer vision problem, where the task is to automatically recognise objects from subordinate categories. 
    One of its main difficulties is capturing the most discriminative inter-class variances among visually similar classes. 
    %
    Recently, methods with Vision Transformer (ViT) have 
    demonstrated noticeable achievements in FGVC, generally by 
    employing the self-attention mechanism with additional resource-consuming techniques to 
    distinguish potentially discriminative regions while disregarding the rest.
    However, such approaches may struggle to effectively focus on truly discriminative regions 
    due to only relying on the inherent self-attention mechanism, 
    resulting in the classification token likely aggregating global information from less-important background patches.
    Moreover, due to the immense lack of the datapoints, classifiers may fail to find the most helpful inter-class distinguishing features, since other unrelated but distinctive background regions may be falsely recognised as being valuable. 
    %
    To this end, we introduce a simple yet effective 
    Salient Mask-Guided Vision Transformer (SM-ViT), where the discriminability of the standard ViT's attention maps is boosted through salient masking of potentially discriminative foreground regions.
    Extensive experiments 
    demonstrate that with the standard training procedure our SM-ViT achieves state-of-the-art performance on popular FGVC benchmarks among existing ViT-based approaches while requiring fewer resources and lower input image resolution.
}

\onecolumn \maketitle \normalsize \setcounter{footnote}{0} \vfill

\section{\uppercase{Introduction}}
\label{sec:introduction}

Fine-grained visual classification (FGVC) is a challenging computer vision task that aims to detect multiple sub-classes of a meta-category (e. g., car or airplane models \cite{6755945,https://doi.org/10.48550/arxiv.1306.5151}, 
animal or plant categories \cite{iNat,flowers}, etc.).
This type of image classification, compared to the traditional one \cite{imagenet_v1,places}, involves larger inter-class similarity and a lack of data per class \cite{9008286}.
This complex yet essential problem has a wide range of research and industrial applications including, but not limited to, autonomous driving, 
visual inspection, object search, and identification.

\begin{figure}[!ht]
  \centering
  \includegraphics[width=0.98\linewidth]{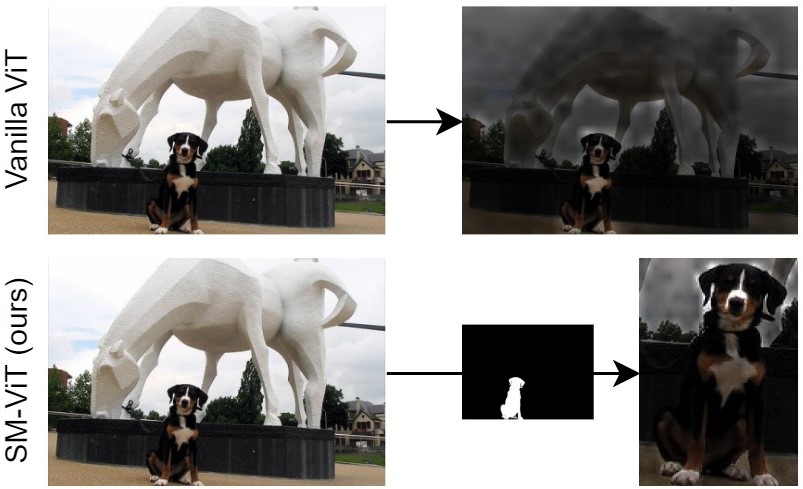}
  \caption{Visualised attention performance comparison of vanilla ViT (first row) and our SM-ViT (second row). 
  For ViT we demonstrate the averaged attention map for the final class token.
  While for SM-ViT we first show the extracted saliency mask from the salient object detection module, and then the final class token's averaged attention map augmented according to this mask. 
  }
  \label{fig:intro}
\end{figure}

One of the important keys to solving the FGVC problem 
is to detect more distinguishable regions in an image \cite{8578534,Luo2019CrossXLF}, and the computer vision community has produced various solutions attempting to do so for the past years \cite{NIPS2011_86b122d4,KHAN201516,ZhengHeliang}. The most recent achievements, based on deep neural networks \cite{7780459}, mainly represent localisation-based and attention-based methods (following \cite{ffvt,TransFG}).
The localisation-related approaches aim to 
learn discriminative and interpretable features from specific regions of input images. Such approaches initially utilised properly annotated parts of an object in each image \cite{6618972,6751314,HuangShaoli}, 
however, impractically laborious and costly densely annotated datasets along with the slow inference of the final model initiated more advanced techniques. Recent works on localisation-based methods \cite{9008286,GeWeifeng}, adopt a region proposal network able to predict regions potentially containing the discriminative features. This information 
is further passed to a backbone 
in order to extract features from these regions \cite{NIPS2009_2f55707d}.
A drawback of such methods is that they are tend to consider the predicted regions as independent 
patches \cite{TransFG}, what may result in inefficiently large bounding boxes simply containing more foreground information than the other potentially more discriminative but smaller proposals. Another issue is that such extra modules often require solving an individual optimisation problem.

Recently, attention-based methods \cite{XiaoTianjun,9578646} leveraging vision transformer (ViT) architecture \cite{vit} have achieved noticeable results on image classification problems. 
ViT considers an input image as a sequence of its patches, what allows the model to aggregate important information from the whole image at a time.
A self-attention mechanism further attempts to detect the most discriminative patches, which help to automatically find the important regions in an image. This way of processing makes the model able to capture long-range dependencies beneficial for classification \cite{ChenJieneng}.
%
Such an ability of the attention-based methods to efficiently learn distinctive features can also be helpful for the FGVC problem as well.
Nevertheless, a recent study \cite{TransFG} investigating the performance of vanilla ViT on FGVC indicates that the class token, deciding on the final class probabilities, may pay more attention to global patches and concentrate less on local ones, which can hamper the performance in fine-grained classification.
Recent ViT-based approaches for FGVC \cite{ffvt,TransFG} typically attempt to solve this issue by introducing an extra module, which is responsible for 
better segregation of class token attention by implicit distinguishing of potentially discriminative regions while disregarding the rest.
However, these methods may struggle to effectively focus on more discriminative regions due to only relying on the self-attention mechanism, resulting in the classification token likely aggregating global information from less valuable background regions.
Moreover, despite the accuracy improvement, such methods 
mainly introduce significantly more computations or trainable parameters. 

\textbf{Contributions:} \textit{(1)} In this work, we introduce a simple yet effective approach to improve the performance of the standard Vision Transformer architecture at FGVC. 
Our method, named Salient Mask-Guided Vision Transformer (SM-ViT), utilises a salient object detection module comprising an off-the-shelf saliency detector to produce a salient mask likely focusing on the potentially discriminative foreground object regions in an image. The saliency mask is then utilised within our ViT-like Salient Mask-Guided Encoder 
(SMGE) to boost the discriminability of the standard self-attention mechanism, thereby focusing on more distinguishable tokens. 
\textit{(2)}~We argue that, in the case of fine-grained classification, the most important features are in the foreground and come from the main (salient) object in an image. However, unlike some of the previous SOTA ViT-based works, we do not completely disregard the less recognisable image parts but rather guide the attention scores towards the more beneficial salient patches.
\textit{(3)}~Moreover, we address the well-known problem of the immense lack of the datapoints in FGVC datasets, when classifiers often fail to find truly helpful inter-class distinguishing features, since unrelated but distinctive background regions may be falsely recognised as being valuable within the little available information provided by a training data set.
Therefore, by encouraging the self-attention mechanism to pay its "attention" to the salient regions, we simply enforce it to concentrate its performance within the main object and, therefore, to find the truly distinguishing cross-class patches.
\textit{(4)}~To the best of our knowledge, we are the first to explore the effective utilisation of saliency masks 
in order to extract more distinguishable information within the ViT encoder layers by boosting the discriminability of self-attention features for the FGVC task.
\textit{(5)} We experimentally demonstrate that the proposed SM-ViT effectively reduces the influence of unnecessary background information while also focusing on more discriminative object regions (see Fig. \nolinebreak \ref{fig:intro}).
%
Our comprehensive analysis of extensive experiments on three popular fine-grained recognition datasets (Stanford Dogs, CUB, and NABirds) demonstrates that with the standard training procedure the proposed SM-ViT achieves state-of-the-art performance on FGVC benchmarks, compared to existing ViT-based approaches published in literature.
\textit{(6)} Another important advantage of our solution is its integrability, since it can be fine-tuned on top of a ViT-based backbone or can be integrated into a Transformer-like architecture that leverages the standard self-attention mechanism.
The code and models are shared at: \href{https://github.com/demidovd98/sm-vit}{\textit{https://github.com/demidovd98/sm-vit}}.
%

\section{\uppercase{Related Work}}

\subsection{Fine-Grained Visual Classification}
Besides the plain feature-encoding CNN-based solutions \cite{https://doi.org/10.48550/arxiv.1807.09915,https://doi.org/10.48550/arxiv.1911.03621,Gao_Han_Wang_Huang_Scott_2020} simply extracting high-order image features for recognition, current specific solutions for the FGVC problem are mainly related to two following groups based on a method used: localisation-based and attention-based approaches. The former aim to explicitly detect discriminative regions and perform classification on top of them, and the latter aim to predict the relationships among image regions and classify the object by this information.

Early localisation-based methods \cite{6618972,HuangShaoli}, initially proposed for densely annotated datasets with bounding boxes for important regions, first locate the foreground object and its parts and then classify the image based on this information. Despite the relatively better performance, such solutions highly rely on the manual dense annotations including one or multiple bounding boxes per image, what makes them practically inapplicable to the real world scenarios.
As a solution for this problem, in \cite{GeWeifeng} the authors leverage weakly-supervised object detection and instance segmentation techniques to first predict multiple coarse regions and further choose the most distinguishable ones.
In later works, \cite{He_Peng_2017} suggested using additional spatial constraints to improve the quality of the chosen parts,
\cite{9156899} presented an approach to utilise potential correlations among parts in order to select the best ones, 
and \cite{https://doi.org/10.48550/arxiv.2102.09875} presented a method able to first create a database of region features and then correct the class prediction by re-ranking the detected global and local information.
Despite the better performance, such approaches usually require a separate, properly constructed, detection branch, which complicates the overall architecture and noticeably increases training and inference time. In addition, complete cropping of less important regions does not always increase the model's accuracy. 

As an alternative, attention-based methods are able to perform both classification and localisation steps simultaneously and with no additional data, by predicting the discriminative parts inside the self-attention mechanism.
For example, in \cite{Zhao_2017} authors propose 
leveraging visual attention to extract different attention maps and find the important information in them.
A multi-level attention technique is presented in \cite{XiaoTianjun}, where the final model is capable of filtering out the common among classes regions.
Later works \cite{https://doi.org/10.48550/arxiv.1807.09915} and \cite{9578646} demonstrate modified architectures with the integration of a cross-layer bilinear pooling mechanism and a progressive attention technique, respectively, where both aim to progressively improve the region prediction performance.
Several other recent solutions for FGVC are \cite{Yu_2021_ICCV,ZHAO2022108229}, which demonstrate different improvements for distinctive regions localisation, and \cite{https://doi.org/10.48550/arxiv.2101.06635,https://doi.org/10.48550/arxiv.2205.02151,DoTuong,https://doi.org/10.48550/arxiv.2203.02751,https://doi.org/10.48550/arxiv.2208.14607}, which propose complex techniques for mainly marginal performance increasing.
Although these methods actually show some improvement, they mostly come with a few noticeable drawbacks:
a significant computational cost or a more complex architecture, resulting in a lack of interpretability and dataset-specific solutions.

\subsection{Vision Transformer}
Initially discovered to process sequences of text in natural language processing (NLP) \cite{https://doi.org/10.48550/arxiv.1706.03762}, the Transformer architecture with its self-attention mechanism has shown a great success in that field \cite{https://doi.org/10.48550/arxiv.1810.04805,https://doi.org/10.48550/arxiv.1901.02860,https://doi.org/10.48550/arxiv.1906.00295} and was later extended by researchers to computer vision (CV) tasks. After the proposal of the Vision Transformer architecture \cite{vit}, which demonstrated the SOTA performance on multiple problems, the community has been gradually exploring ViT's abilities by using it as a backbone for popular CV problems, such as image classification \cite{https://doi.org/10.48550/arxiv.2112.13492}, object detection \cite{CarionNicolas,https://doi.org/10.48550/arxiv.2010.04159}, segmentation \cite{XieEnze,https://doi.org/10.48550/arxiv.2012.00759,9578646} and others \cite{GirldharRohit,https://doi.org/10.48550/arxiv.2012.15460,https://doi.org/10.48550/arxiv.2102.04378}. Simple integration of the ViT architecture into other backbones and techniques has gained promising achievements and still remains SOTA for various problems.
However, only a few studies investigate the properties of Vision Transformer on the FGVC problem, where the vanilla ViT model shows worse performance than its CNN counterparts. 
One of the pioneers leveraging ViT on FGVC tasks is TransFG framework \cite{TransFG}, which is the first to propose a solution to automatically select the distinguishable image patches and later use them for the final classification step. However, in order to achieve better results, this method uses overlapping patches, what requires significantly more resources, compared to vanilla ViT.
Further proposed 
FFVT \cite{ffvt} uses its special MAWS module for feature fusion, what makes it able to aggregate more local information from the ViT encoder layers, what, as the authors stated, improves the original ViT feature representation capability. However, its overall idea includes selecting the patches with the most attention scores and then disregarding the other ones. This concept, based on the imperfect self-attention mechanism, may increase the negative for FGVC effect of background patches. 

\subsection{Mask-guided Attention}
Similar to our work, recently proposed mask-guided attention techniques, mostly based on primitive saliency models, have also demonstrated promising results in detection and re-identification tasks \cite{9282190}.
However, only few studies have attempted to explore its ability to be helpful for fine-grained visual classification. For example, in \cite{8578227} the authors suggest adding mask information as an additional input channel for a CNN in order to separately learn features from the original image and both foreground and background masked copies of it.
Another work, \cite{Wang_2021}, investigates the capacity of such models when trained on more difficult patchy datasets, where the authors offer to use the predicted mask to guide feature learning in the middle-level backbone layers.
In addition to that, some other early writings also discuss the ways of leveraging the saliency information to guide the learning process by simply using the mask as an extra input \cite{8354258,HagiwaraAiko}.
Another close to our idea recent approach is \cite{ijcai2022p138}, where the authors suggested using convolution kernels with different sizes for input patches. These feature activations are embedded into a ViT-like encoder in order to increase its locality and translation equivariance qualities for binary medical image segmentation problems. However, the authors do not directly use the actual saliency information, and rather assume that different activation maps are coming from different types of the same single positive class. 
Recently, several works \cite{https://doi.org/10.48550/arxiv.1712.06492,https://doi.org/10.48550/arxiv.2008.05413} suggested integration of saliency prediction as an additional component in a loss function in order to improve the training procedure. This is achieved by feeding both the image and the predicted target saliency map, so that the model is trained to produce outputs similar to the map.
Another idea of leveraging saliency information in attention-based models is described in \cite{Yu_Huang_Wang_Cheng_Chu_Cui_2022}. The authors suggested using attention maps from each encoder layer to generate a saliency binary mask, which is then used for model pruning.
All these methods indeed demonstrate that the fusion of saliency information with the main architecture's data flow can be beneficial and efficient, however they do not directly utilise the saliency and rather use it as collateral data, what results in a less noticeable 
improvement.  
%

\section{\uppercase{Method}}

\begin{figure*}[!ht]
  \centering
  \includegraphics[width=0.85\linewidth]{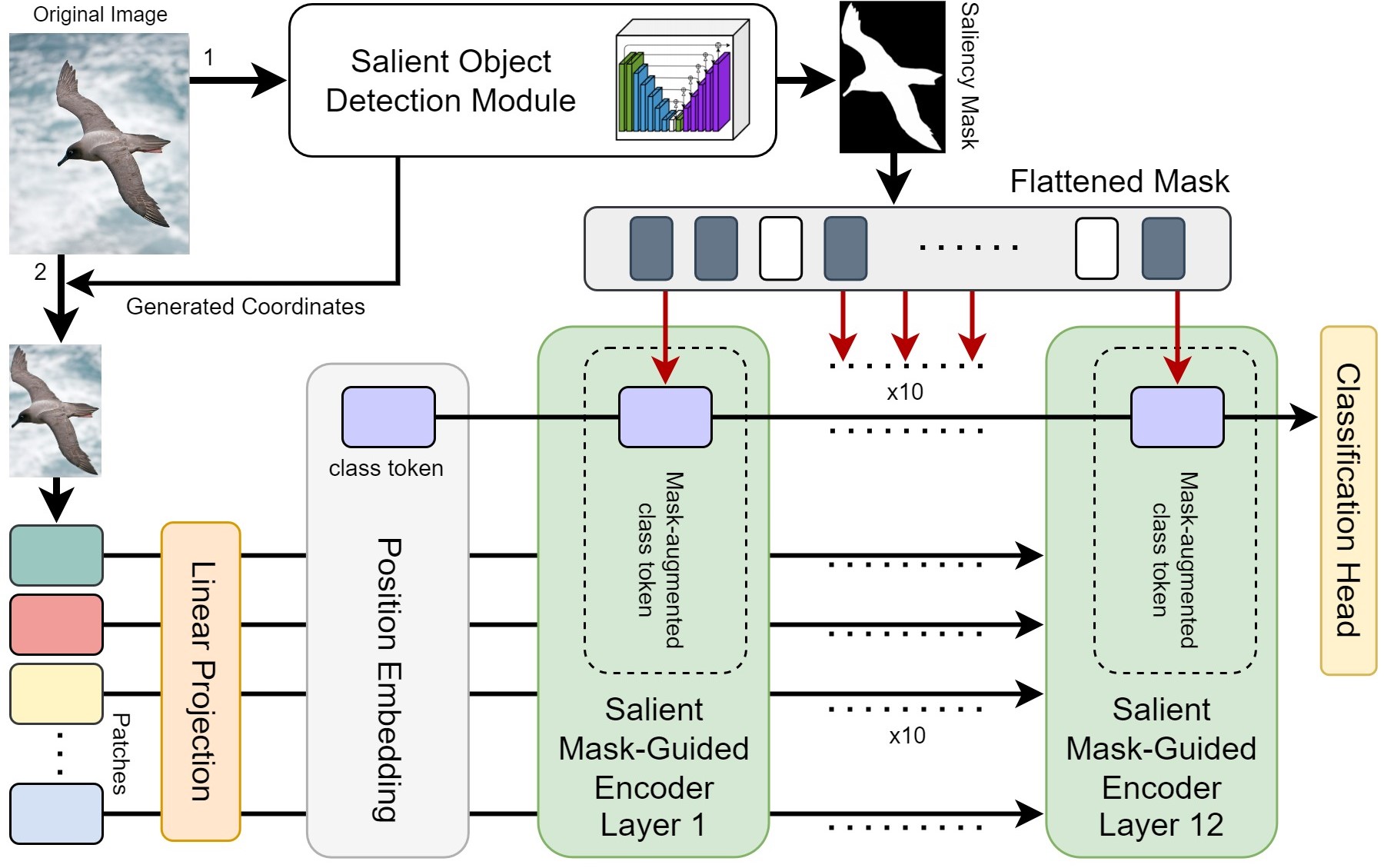}
  \caption{The overall architecture of our proposed SM-ViT. 
  An image is first fed into the salient object detection module to extract its saliency mask and automatically generate a bounding box, which are then used to prepare a binary mask and to crop the image respectively.
  Further, the cropped image is fed into the ViT-like architecture, where it is first split into patches, projected into the embedding space, the positional embedding is added to the patches, and a class token is prepended.
  Next, the resulted sequence of tokens is passed through each layer of our Salient Mask-Guided Encoder (SMGE), where inside the multi-head self-attention mechanism the flatten binary mask is used to augment attention scores of the class token accordingly.
  Lastly, the class token values from the last SMGE layer are passed to a classification head to perform categorisation.}
  \label{fig:general}
\end{figure*}

\subsection{Vision Transformer Framework}

Following the original Transformer \cite{https://doi.org/10.48550/arxiv.1706.03762} made for NLP tasks, the standard Vision Transformer architecture \cite{vit} also expects an input to be a 1D sequence of tokens. Therefore, in order to adapt it to computer vision problems, 2D input images need to be first cropped into smaller 2D patches and then flattened into 1D vectors, so that the input dimension changes are following: $(H, W, C) \rightarrow (N_p, P, P, C) \rightarrow (N_p, (P^2 * C))$, where $H, W, C$ are initial image sizes, $P$ is a predefined patch size, and $N_p = H*W/P^2$ is the number of such patches.
Next, using a trainable linear projection, transformed patches are mapped into a latent embedding space of dimension $D$, which is the vector size of all tokens throughout all ViT layers.
Learnable 1D position embeddings are further added to the patches in order to preserve the information about spatial relations among them. 
Therefore, the resulting token embedding procedure is as follows:
\begin{equation}
    \label{eq:patch_emb}
    \textbf{z}_0 = [x_{cls}; x_p^1 \textbf{E}; x_{p}^{2} \textbf{E}; ... ;  x_{p}^{N_p} \textbf{E}] +\textbf{E}_{pos},
\end{equation}
where 
$\textbf{E} \in \mathbb{R}^{(P^2 * C) \times D}$ is the trainable linear projection and $\textbf{E}_{pos} \in \mathbb{R}^{(N+1) \times D}$ is the position embedding.

As the last input preparation step, an extra learnable class token is pre-pended to the sequence, so that it can interact with the image patches, similar to \cite{https://doi.org/10.48550/arxiv.1810.04805}.
This token, fed to the encoder with the sequence of embedded patches, is supposed to aggregate the information from the image tokens in order to summarise the image representation and convey it to a classification head.

In more detail, the vanilla ViT encoder component, same as in \cite{https://doi.org/10.48550/arxiv.1706.03762}, consists of several repeating encoding layers utilising the multi-head self-attention (MSA) mechanism, MLP blocks, and both layer normalisation (LN) and residual connection techniques \cite{https://doi.org/10.48550/arxiv.1809.10853,https://doi.org/10.48550/arxiv.2012.00759}.
More specifically, MSA used in vanilla ViT is an extension of the ordinary self-attention mechanism, represented by this equation:
\begin{equation}
    \label{eq:attention}
    Attention(Q,K,V) = softmax( \frac{Q K^T} {\sqrt{d_k}} ) V
\end{equation}
where $d_k$ is a scaling factor equal to the dimension number of $Q,K,V$, which are queries, keys, and values respectively, derived from the input patches.

Eventually, the classification head, implemented as a multi-layer perceptron (MLP) with a hidden layer, is attached to the class token $\textbf{z}_L^0$ in the last encoder layer $L$ and is responsible for the final category prediction, based on the aggregated information.

\subsection{Salient Mask-Guided ViT}

\textbf{Overall Architecture.}
The ViT architecture, initially designed for less fine-grained problems, is supposed to capture both global and local information, what makes it spend a noticeable part of its attention performance on the background patches \cite{vit}. This property makes vanilla ViT perform worse on FGVC tasks, since they usually require finding the most distinguishable patches, which are mostly the foreground ones.
In order to resolve this issue, we propose Salient Mask-Guided Vision Transformer (SM-ViT), which is able to embed information coming from a saliency detector into the self-attention mechanism.
The overall architecture of our SM-ViT is illustrated in Fig. \ref{fig:general}.

\textbf{Salient Object Detection Module.}
At the initial step, we utilise a salient object detection (SOD) module for saliency extraction. Our proposed method employs a popular deep saliency model, U2-Net \cite{Qin_2020}, pre-trained on a mid-scale dataset for salient object detection \cite{Wang_2017_CVPR}. 
We chose this particular solution since its nested U-shaped architecture predicts saliency based on rich multi-scale features at relatively low computation and memory costs.
First, an input image is passed through the 
SOD module set up in a test mode, which further generates 
the final non-binary saliency probability map.
In the next phase, the model output is normalised to be within the values $[0 ... 1]$ and then converted into a binary mask by applying a threshold $d_{\alpha}$ on each mask's pixel $a_i$ according to the following equation:
\begin{equation}
    \label{eq:threshold}
    a_i = 
    \begin{cases}
    1,  & \text{if } a_i \geq d_{\alpha} \\
    0,  & \text{otherwise,}
    \end{cases}
\end{equation}
where $d_{\alpha}$ is a pixel's intensity threshold.
According to the authors' recommendations \cite{Qin_2020}, we set it to $0.8$ for all our experiments. 
Finally, the resulting binary mask 
and a bounding box for the found salient object(s) (in the form of the minimum and maximum 2D coordinates of the positively thresholded pixels) are extracted and saved. 
An important note is that our solution also takes into account the cases when a mask is not found or is corrupted, and, if so, the initial probability map is first refined again with Eq. (\ref{eq:threshold}) using a threshold $d_{\alpha}=0.2$, which allows more pixels to be considered positive.
If the mask is not restored even after refining, its values are automatically set as positive for the central 80 \% of the image pixels.
The extracted binary mask and bounding box are further passed into our SMGE.

\textbf{Salient Mask-Guided Encoder.}
The core module of SM-ViT is our novel Salient Mask-Guided Encoder (SMGE), which is a ViT-like encoder modified to be able to receive and process saliency information. Its main purpose is to increase the class token attention scores for the image tokens containing foreground regions.
Initially, an image, cropped according to the extracted in SOD module bounding box, in a form of patches is projected into linear embeddings, and a position embedding is added to it.
Next, instead of the standard ViT's encoder, our SMGE takes its place functioning as an improved self-attention mechanism.
In order to understand the intuition behind our idea, we need to point out that the way of attention obtaining in the vanilla ViT encoder (refer to Eq \ref{eq:attention}) makes the background and foreground patches equally important and does not discriminate valuable for FGVC problems salient regions of the main object(s) in an image.
Taking this issue into account, our solution is to increase attention scores for the patches that include a part of the main (salient) object in them.
However, due to the nature of the self-attention mechanism and the non-linearity used in it, one can not simply increase the final attention values themselves, since it will break the major assumptions of the algorithm \cite{vit}.
Therefore, in order to solve this problem, we apply changes to attention scores calculated right before the softargmax function (also known as softmax), according to the saliency mask provided by the salient object detection module.
For this purpose, the binary mask is first flattened into a 1D vector and a value for the class token is pre-pended to it, so that, similar to Eq \ref{eq:patch_emb}, the size of the resulting mask matches the number of tokens ($N_{p}+1$):
\begin{equation}
    \label{eq:mask}
    \textbf{m} = [m_{cls}; m_{p}^1; m_{p}^{2}; ... ;  m_{p}^{N_p}],
\end{equation}
where $m_{cls}$ is always positive since the attention of the class token to itself is considered favourable \cite{vit}.
Further, a conventional attention scores matrix $X_{scor}$ is calculated in each head:
\begin{equation}
    \label{eq:SAM_1}
    X_{scor} = \frac{Q K^T} {\sqrt{d_k}} V
\end{equation}
Next, the maximum value $\textbf{x}_{max}$ among the attention scores of the class token to each patch is found for each head.
These values are further used to modify the attention scores of the class token by increasing the unmasked by $\textbf{m}$ ones with a portion of the largest found value $\textbf{x}_{max}$, what is calculated for every head:
\begin{equation}
    \label{eq:SAM_2}
    \textbf{x}_{scor_{cls}} = 
    \begin{cases}
    x_{scor_{cls}}^{i} + ( x_{max} * d_{\theta})  & \text{if } m_i = 1 \\
    x_{scor_{cls}}^{i}  & \text{otherwise,}
    \end{cases}
\end{equation}
where $\textbf{x}_{scor_{cls}}$ is a row in the matrix of attention scores $X_{scor}$ belonging to the class token, $d_{\theta}$ is a coefficient controlling the portion of the maximum value to be added, and $i \in [1, 2, ... N_p, N_{p}+1]$. We provide an ablation study on the choice of coefficient $d_{\theta}$ in Section \nolinebreak \ref{sec:ablation}.
Finally, following Eq \ref{eq:attention}, the rows of the resulted attention scores matrix $X'_{scor}$, 
including the modified values in its $\textbf{x}_{scor_{cls}}$ row, are converted into probability distributions using a non-linear function:
\begin{equation}
    \label{eq:SAM_3}
    Y = softmax(X_{scor})
\end{equation}
Eventually, similar to the multi-layer vanilla ViT encoder, the presented algorithm is further repeated at each SMGE's layer until the classification head, where the standard final categorisation is done based on the class token aggregating the information from the "highlighted" regions throughout SMGE.
To summarise, our simple yet efficient salient mask-guided encoder changes the vanilla ViT encoder by modifying its standard attention mechanism's algorithm (in \nolinebreak Eq \nolinebreak \ref{eq:attention}) with Eq \ref{eq:mask}-\ref{eq:SAM_3}. Therefore, relatively to the vanilla ViT encoder, our SMGE only adds pure mathematical steps, does not require extra training parameters, and is not resource costly.

\section{\uppercase{Experiments and Results}}

In this section, we describe in detail the setup used for our experiments, compare the obtained results with current state-of-the-art achievements, and provide an ablation study containing quantitative and qualitative analysis. 
We demonstrate and explore the ability of our SM-ViT to utilise saliency information in order to improve its performance on FGVC problems.

\subsection{Experiments Setup}

\begin{table}[!ht]
  \caption{The details of three fine-grained visual classification datasets used for the experiments.}
  \label{tab:datasets}
  \centering
  \begin{tabular}{c|c|c|c}
    \toprule
    Dataset & Categories & Classes & Images  \\
   
    \midrule
    
    Stanford Dogs & Dogs & 120 & 20,580 \\
    
    CUB-200-2011 & Birds & 200 & 11,788 \\
    
    NABirds & Birds & 555 & 48,562 \\
  \bottomrule
\end{tabular}
\end{table}

\begin{table*}[!ht]
  \caption{Accuracy comparison on three FGVC datasets for our SM-ViT and other SOTA ViT-based methods.
  The considered methods use the ViT-B/16 model pre-trained on the ImageNet-21K dataset. 
  All of them are then fine-tuned with the standard ViT training procedure with no overlapping patches. The best accuracies are highlighted in bold.
  }
  \label{tab:sota}
  \centering
  \begin{tabular}{c|c|c|c|c}
    \toprule
    Method & Backbone & Stanford Dogs & CUB-200-2011 & NABirds \\
   
    \midrule
    ViT \cite{vit} & ViT-B/16 & 91.4 & 90.6 & 89.6 \\
    
    TPSKG \cite{LiuXinda} & ViT-B/16 & 91.8 & 91.0 & 89.9 \\
    
    DCAL \cite{https://doi.org/10.48550/arxiv.2205.02151} & ViT-B/16 & - & 91.4 & - \\ 
    
    TransFG \cite{TransFG} & ViT-B/16 & 91.9 & 91.5 & 90.3 \\ 

    SIM-Trans \cite{https://doi.org/10.48550/arxiv.2208.14607} & ViT-B/16 & - & 91.5 & - \\ 

    AFTrans \cite{https://doi.org/10.48550/arxiv.2110.01240} & ViT-B/16 & 91.6 & 91.5 & - \\

    FFVT \cite{ffvt} & ViT-B/16 & 91.5 & \textbf{91.6} & 90.1 \\

    \midrule
    SM-ViT (Ours) & ViT-B/16 & \textbf{92.3} & \textbf{91.6} & \textbf{90.5} \\
    
    \bottomrule
\end{tabular}
\end{table*}

\textbf{Datasets.}
We explore the properties of our SM-ViT on three different popular benchmarks for FGVC: Stanford Dogs \cite{KhoslaYao}, CUB-200-2011 \cite{WelinderPeter}, and NABirds \cite{Horn2015} (for more details, see Table \ref{tab:datasets}).
From the chosen datasets, Stanford Dogs and CUB-200-2011 are considered medium-sized FGVC benchmarks, and NABirds is a large-sized one.
We also emphasise that, despite its size, the Stanford Dogs dataset includes images with multiple objects, artificial objects, and people. It makes the task harder for the saliency extraction module due to the pre-training objective's shift towards humans.

\textbf{Baselines and Implementation details.}
%
For all our experiments, the backbone for the classification part is Vision Transformer, more specifically, a ViT-B/16 model pre-trained on the ImageNet-21K \cite{imagenet_v1} dataset with 224x224 images and with no overlapping patches.
For the saliency detection module, a U2-Net model pre-trained on the DUTS-TR \cite{Wang_2017_CVPR} dataset, is used with constant weights. 
Following common data augmentation techniques, unless stated otherwise, the image processing procedure is as follows.
For the saliency module, as recommended by the authors \cite{Qin_2020}, the input images are resized to 320x320 and no other augmentations are applied.
For our SM-ViT, the images are resized to 400x400 for the Stanford Dogs and CUB-200-2011 datasets and to 448x448 for NABirds (due to its higher-resolution images), without cropping for all datasets.
Next, only for the training process, random horisontal flipping and colour jittering techniques are applied.
The threshold $d_{\theta}$ in Eq. (\ref{eq:SAM_2}) is set to 0.25 for CUB and NABirds and to 0.3 for Stanford Dogs (see Section \ref{sec:ablation} for ablation details).
All our models are trained with the standard SGD optimiser with a momentum set to 0.9 and with a learning rate of 0.03 for CUB and NABirds, and 0.003 for Stanford Dogs, all with cosine annealing for the optimiser scheduler.
The batch size is set to 32 for all datasets.
Pre-trained with 224x224 images ViT-B/16 weights are load from the official ViT \cite{vit} resources.
For a fair comparison, we also reimplement some of the methods with the above-mentioned preset while also following their default settings. 
All our experiments are conducted on a single NVIDIA RTX 6000 GPU using the PyTorch deep learning framework and the APEX utility. 

\begin{figure*}[!ht]
  \centering
  \includegraphics[width=0.98\linewidth]{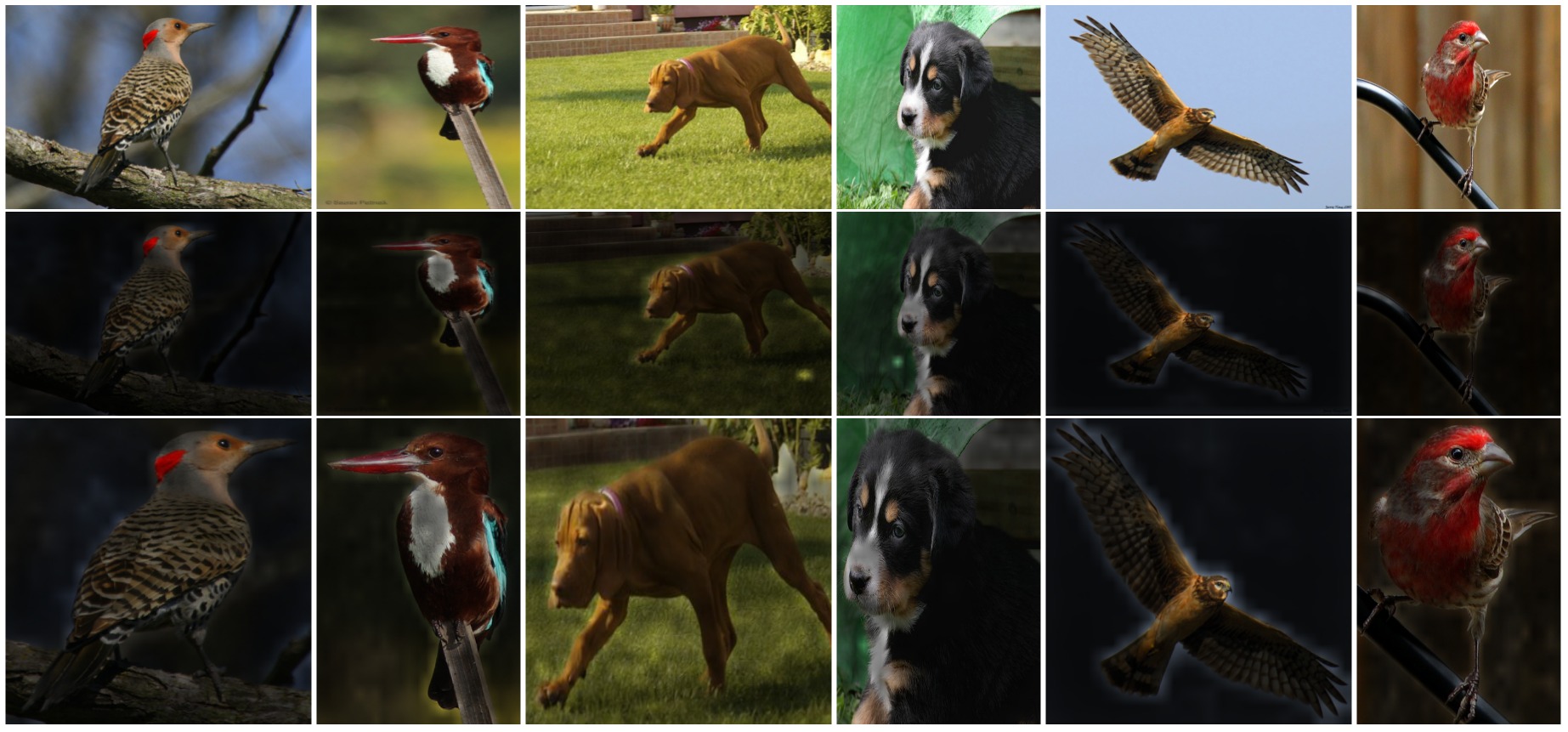}
  \caption{Visualisation of vanilla ViT and our SM-ViT results on different datasets. The first row shows original images, while the second and third rows demonstrate averaged by all heads attention maps generated by the class token at the final encoder layer of vanilla ViT and SM-ViT, respectively. Brightness intensity represents the total amount of attention, where the more attention the class token pays to a region, the brighter it is, and the other way around.
  }
  \label{fig:ablation}
\end{figure*}

\subsection{Comparison With State-Of-The-Art}
In this subsection, we compare the performance of our SM-ViT with other ViT-based SOTA methods on three FGVC datasets.
Before discussing the results, we need to emphasise that our initial goal is to provide an improved Vision Transformer architecture which is able to perform on FGVC problems better than the original ViT and also can easily replace it in other works where it is used as a backbone. We are designing an approach to improve the baseline without additional training parameters and significant architecture changes, rather than simply providing the best-performing but hardly applicable solution "by all means".
Keeping this in mind, in Table \nolinebreak \ref{tab:sota} we compare our SM-ViT with other officially published SOTA ViT-based approaches, which only use the ViT-B/16 backbone with no significant changes and do not require a lot of extra computations or training parameters compared to vanilla ViT. There also exist other methods which mainly use significantly more complex solutions, either requiring noticeably more training time and resources, or using more sophisticated and less popular backbones.

The results on Stanford Dogs demonstrate that besides a significant improvement of 0.9 \% over vanilla ViT, our method is also superior among other approaches utilising the unchanged ViT-B/16 backbone showing a margin of 0.4 \% to the second best counterpart.
It is important to mention that images in this dataset often include multiple extraneous objects (e.g. other animals, multiple categories, artificial objects) besides the main category, 
which may negatively affect the performance of the salient object detection module. Nevertheless, our SM-ViT still manages to demonstrate the best result.
%
On CUB our solution outperforms vanilla ViT by noticeable 1.0 \nolinebreak \% and also shares the Top-1 performance with another ViT-based solution, FFVT, which shows noticeably lower performance on the other datasets.
%
For NABirds, our method improves vanilla ViT performance by up to 90.5 \%, showing a margin of 0.9 \% over the predecessor and of 0.2 \% over the closest counterpart.
We also emphasise that, our approach provides almost equally large performance increase on each considered dataset, what certainly demonstrates its ability to adapt to a problem and generalise better. We point out that, unlike some of the considered dataset-specific methods, our goal was not the proposal of an over-optimised on a singular dataset solution, but rather a potentially widely-applicable automatic approach with fewer heuristic parameters to adjust.

\subsection{Ablation Studies}
\label{sec:ablation}

\subsubsection{Effect of $d_{\theta}$ coefficient}

With the goal to investigate the influence of the heuristic part of our solution, we provide an ablation study on the effect of different values for hyper-parameter $d_{\theta}$ in Eq \nolinebreak \ref{eq:SAM_2}. 
The results of the experiments can be found in Table \nolinebreak \ref{tab:coeff}.
Therefore, following the ablation results, the best $d_{\theta}$ value for CUB and NABirds is 0.25, and for Stanford Dogs is 0.3.
One can also observe that performance is better with this coefficient value within the $0.2-0.3$ range, so we can suggest that for other FGVC datasets this range can be a good starting point.
We assume that this is the case due to the nature of the self-attention mechanism. Making the attention scores too large compared to other patches makes it too discriminative compared to the less distinguishable and background regions, since the final difference among output values grows exponentially.
In addition, one know that the attention mechanism is not perfect and can not always identify the most important regions, let alone the fact that the performance of saliency detectors is also imperfect so they may produce messy predictions.
Moreover, in some cases some information about background can be especially helpful 
so it may be unnecessary to completely crop out the background patches.

\begin{table}[!ht]
  \caption{
  The effect of different values for hyper-parameter $d_{\theta}$ in \nolinebreak Eq \nolinebreak \ref{eq:SAM_2}. Training procedure and the rest hyper-parameters remain unchanged. The best performance is highlighted in bold.}
  \label{tab:coeff}
  \centering
  \begin{tabular}{c|c|c|c}
    \toprule
    Value of $d_{\theta}$ & Stanford Dogs & CUB & NABirds  \\
   
    \midrule
    0.1 & 92.1 & 91.4 & 90.3 \\
    \hline
    
    0.25 & 92.2 & \textbf{91.6} & \textbf{90.5} \\
    \hline

    0.3 & \textbf{92.3} & 91.5 & 90.4 \\
    \hline

    0.5 & 92.1 & 91.3 & 90.2 \\
    \hline

    1.0 & 92.0 & 91.1 & 90.1 \\
  \bottomrule
\end{tabular}
\end{table}

All these points make the choice of $d_{\theta}$ a trade-off between the performance of the utilised saliency extractor and attention-based backbone.

\begin{table*}[!ht]
  \caption{The effect of our SMGE method, applied at different stages in our SM-ViT. The indicated performance is for the CUB-200-2011 dataset with the standard ViT training and validation procedure. For inference experiments, the results represent the average time per image in a batch size of 16.
  }
  \label{tab:crop}
  \centering
  \begin{tabular}{c|c|c|c|c|c}
    \toprule
    Method & Img. Resolution & SMGE           &  SMGE         & Accuracy   & Inference time,  \\
        &    & in training  &  in inference &       &  relative increase \\
    \midrule
    
    ViT & 448x448 & $\times$ & $\times$ & 90.6 & x1.0 \\ 
    \hline
    
    ViT + SMGE & 448x448 & $\times$ & \checkmark & 90.8 & x2.2 \\ 
    \hline
    
    ViT + SMGE & 400x400 & \checkmark & $\times$ & 91.1 & x0.6 \\ 
    \hline

    SM-ViT (Ours) & 400x400 & \checkmark & \checkmark & 91.6 & x1.4 \\ 

  \bottomrule
\end{tabular}
\end{table*}

\subsubsection{Effect of SMGE}
\begin{figure}[!ht]
  \centering
  \includegraphics[width=0.9\linewidth]{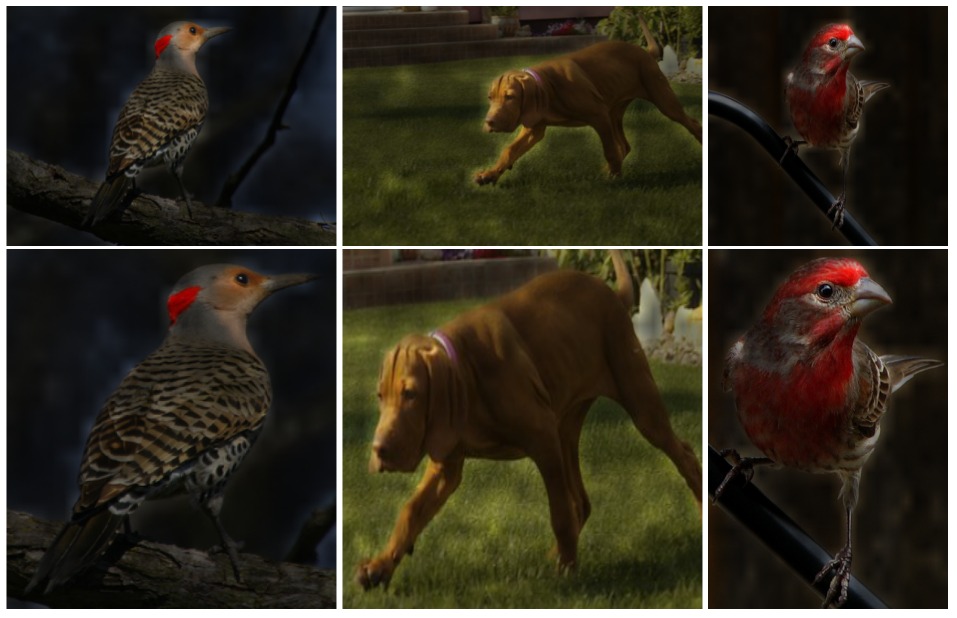}
  \caption{Visual comparison of class token averaged attention maps at the last encoder layer of our SM-ViT, first fine-tuned with SMGE, and then used with disabled (first row) and enabled (second row) SMGE during inference.}
  \label{fig:crop_mask}
\end{figure}

In Table \nolinebreak \ref{tab:crop} we provide the results of our SM-ViT with our SMGE module at different stages to prove that although mask cropping with the saliency mask automatically generated by the SOD module indeed improves performance, the main effect is achieved mostly by our SM-ViT method altogether. We can see that even with the SMGE module applied during training only, its ability to effectively embed saliency information into the self-attention mechanism allows the model to still have better accuracy than the original ViT has.
For better understanding of this idea, the visualised outputs of two different SMGE setups are provided in Figure \ref{fig:crop_mask} for comparison. We can observe that SM-ViT, trained with SMGE and then utilised without SMGE during inference, still produces good-quality attention maps even without the helpful cropping and explicit attention scores augmentation techniques. It can be seen that in both cases the attention maps generated for the same images by SM-ViT are more pronounced and cover more potentially discriminative regions, compared to vanilla ViT.

It is also worth mentioning that our solution not only outperforms (or is on par with) other SOTA ViT-based methods but also does it with lower resolution inputs. In addition, according to the ablation results in Table \nolinebreak \ref{tab:crop}, our SM-ViT is faster and still more accurate than vanilla ViT when used with SMGE disabled for inference. It becomes possible due to the fact that our solution utilises saliency masks automatically generated by the SOD module. These binary masks are used to crop the main salient object in an image before it is processed by SMGE. This noticeably reduces the original input images and, therefore, allows the model to perform well enough with lower resolutions.
More specifically, our SM-ViT achieves SOTA results with images of 400x400 resolution, compared to its counterparts that require at least 448x448 inputs to achieve a lower or comparable performance. In particular, while achieving better than vanilla ViT results, our SM-ViT requires 15 \nolinebreak \% fewer computations for trainable parameters when using SMGE and 40\% less inference time when it is disabled for inference, which are significant benefits for a resource-intensive ViT-based module.
%

\begin{table}[!ht]
  \caption{
  The effect of different components of our SMGE on the overall performance. The indicated accuracies are for the CUB-200-2011 dataset.}
  \label{tab:smge}
  \centering
  \begin{tabular}{c|c|c|c}
    \toprule
    Method & Saliency & Guided & Acc.  \\
     & Cropping & Attention & \\
    \midrule
    
    ViT & $\times$ & $\times$ & 90.6 \\
    \hline
    
    ViT + SMGE & \checkmark & $\times$ & 90.9 \\
    \hline

    SM-ViT (Ours) & \checkmark & \checkmark & 91.6 \\

  \bottomrule
\end{tabular}
\end{table}

In order to understand the influence of our applied techniques inside SMGE, we provide a performance comparison with different setups in Table \ref{tab:smge}. Based on the results, we emphasise that although mask cropping actually helps to improve the accuracy, it is not the main reason for the performance increase.

\subsubsection{Qualitative Analysis}
To better understand the significance of our SM-ViT, we demonstrate its real outputs and compare them with vanilla ViT ones using images from different datasets. 
The comparison is shown in Fig \ref{fig:ablation}, where the first row contains input images, while the second and third rows demonstrate class token attention maps of vanilla ViT and our SM-ViT, respectively. The attention maps are obtained by averaging the weights of all heads at the last encoder layer, and brightness represents the amount of attention, where the brighter the regions are, the more attention the class token pays to it, and the other way around.
From the results, it is noticeable that SM-ViT provides more focused on the salient object attention maps, which cover more diverse and potentially discriminative parts of the main objects (e. g., colourful feathering, beak, wings shapes, and distinguishable colour patterns). Such an ability also makes the category predictions more robust to natural augmentations and also allows the class token to detect more visually distinctive parts at the same time.
From our extensive qualitative analysis of visualised attention maps obtained by random images from the datasets, we noticed that SM-ViT mostly predicts either better or similar to vanilla ViT attention maps, and only rarely it highlights visually less discriminative regions.

\section{\uppercase{Conclusion}}

In this work, we propose a novel SM-ViT method able to improve the performance of vanilla Vision Transformer on FGVC tasks by guiding the attention maps towards potentially more important foreground objects and, therefore, reducing its "spreading" to less distinguishable regions.
Our core, simple yet efficient salient mask-guided encoder boosts attention efficiency by simply utilising saliency information, does not require additional training parameters, and is relatively not resource costly.
Experimental results demonstrate that with a comparable amount of resources, our SM-ViT is able to produce better than (or similar to) SOTA results among other ViT-based approaches while remaining noticeably efficient.
Based on the promising results, we expect our solution to improve performance on other FGVC datasets containing classes naturally similar to the ones used for saliency module pre-training.
Moreover, we emphasise that our proposed SMGE 
can be further extended to other popular ViT-like backbones with the conventional self-attention mechanism (e.g., DeiT \cite{https://doi.org/10.48550/arxiv.2012.12877}, Swin-T \cite{https://doi.org/10.48550/arxiv.2103.14030}). In addition, other, more powerful salient object detectors producing standard saliency maps, can be used.
Therefore, we believe that SM-ViT has great potential to further boost the performance of various FGVC setups and could be a good starting point for future work.

\bibliographystyle{apalike}
{\small
\bibliography{bib}}

\end{document}